\documentclass[letterpaper, 10 pt, conference]{ieeeconf}  % Comment this line out if you need a4paper

\IEEEoverridecommandlockouts                              % This command is only needed if 
\usepackage{graphics} % for pdf, bitmapped graphics files
\usepackage{epsfig} % for postscript graphics files
\usepackage{times} % assumes new font selection scheme installed
\usepackage{amsmath} % assumes amsmath package installed
\usepackage{amssymb}  % assumes amsmath package installed
\usepackage{cite}
\usepackage{array}
\usepackage{float}
\usepackage{multirow}
\usepackage{booktabs}
\usepackage{subfigure}
\usepackage{algorithm}  
\usepackage{algorithmic}
\newcommand{\argmax}{\mathop{\mathrm{argmax}}\limits}

\title{\LARGE \bf
Interaction-aware Multi-agent Tracking and Probabilistic Behavior Prediction via Adversarial Learning 
}

\author{Jiachen Li*, Hengbo Ma* and Masayoshi Tomizuka
    \thanks{*The authors contributed equally to this work.}
	\thanks{J. Li, H. Ma and M. Tomizuka are with the Department of Mechanical Engineering, 
		University of California, Berkeley, CA 94720, USA
		(e-mail: {\tt\small jiachen\_li, hengbo\_ma, tomizuka@berkeley.edu})}}

\begin{document}

\maketitle
\thispagestyle{empty}
\pagestyle{empty}

%%%%%%%%%%%%%%%%%%%%%%%%%%%%%%%%%%%%%%%%%%%%%%%%%%%%%%%%%%%%%%%%%%%%%%%%%%%%%%%%
\begin{abstract}
In order to enable high-quality decision making and motion planning of intelligent systems such as robotics and autonomous vehicles, accurate probabilistic predictions for surrounding interactive objects is a crucial prerequisite.
Although many research studies have been devoted to making predictions on a single entity, it remains an open challenge to forecast future behaviors for multiple interactive agents simultaneously.
In this work, we take advantage of the Generative Adversarial Network (GAN) due to its capability of distribution learning and propose a generic multi-agent probabilistic prediction and tracking framework which takes the interactions among multiple entities into account, in which all the entities are treated as a whole.
However, since GAN is very hard to train, we make an empirical research and present the relationship between training performance and hyperparameter values with a numerical case study. The results imply that the proposed model can capture both the mean, variance and multi-modalities of the groundtruth distribution.
Moreover, we apply the proposed approach to a real-world task of vehicle behavior prediction to demonstrate its effectiveness and accuracy. 
The results illustrate that the proposed model trained by adversarial learning can achieve a better prediction performance than other state-of-the-art models trained by traditional supervised learning which maximizes the data likelihood.
The well-trained model can also be utilized as an implicit proposal distribution for particle filtered based Bayesian state estimation.
\end{abstract}

%%%%%%%%%%%%%%%%%%%%%%%%%%%%%%%%%%%%%%%%%%%%%%%%%%%%%%%%%%%%%%%%%%%%%%%%%%%%%%%%
\section{INTRODUCTION}
Accurate real-time state estimation and behavior prediction of surrounding objects play a crucial role in intelligent systems (e.g., industrial robotics and autonomous vehicles) since they serve as a premise for appropriate decision making, trajectory planning and motion control. 
Traditional approaches based on state space dynamic systems such as extended Kalman filter or based on time-series analysis work well in simple scenarios where the tracked or predicted targets behave independently.
However, in highly interactive situations it is desired to take into account the inherent uncertainty of future and interactions among different entities since the future behavior of one agent depends heavily on others' behaviors. Moreover, even with the same environment setting, the agent may make different decisions and behave diversely. 
In order to capture these characters of prediction, learning based models with large capacity are necessary.

There have been numerous works focusing on tracking and forecasting future behaviors and motions of humans considering mutual reactions. 
In \cite{human1,human2,tracking4,social_LSTM}, the social behaviors among a group of pedestrians in crowded contexts were modeled by deep neural networks.
Many recent studies have also been devoted to predicting driver behaviors or future trajectories of autonomous vehicles with various models such as intelligent driver model \cite{prediction8}, multi-layer perceptron \cite{prediction4} and recurrent neural network \cite{prediction2,prediction6} which are deterministic models as well as mixture model \cite{prediction1,prediction7,jiachen_prediction}, probabilistic graphical models \cite{jiachen_prediction,prediction4,wei_metric} and variational auto-encoder \cite{prediction3} which are probabilistic models.
There are also studies combining adversarial learning with behavior cloning or reinforcement learning.
In \cite{GAIL1}, Alex et al. extended Generative Adversarial Imitation Learning (GAIL) framework proposed in \cite{GAIL2} to train an action policy network for autonomous vehicles in which only the actions of the ego vehicle are modeled. 
Instead of modeling each agent individually like above studies, in this work we treat multiple interactive agents as a whole system and model the joint distribution of their future behaviors.
Moreover, unlike \cite{GAIL1} where the state-action trajectories were modeled as the output of generator, our method models the distribution of actions at current time step given a sequence of historical states.

The contributions of this work are summarized as follows:
First, we propose a generic framework of adversarial learning which is utilized for probabilistic prediction of multi-agent behaviors. The framework can be generalized to any time-series prediction tasks. 
Second, we make an empirical study to find the relationship between the regularization coefficient of consensus optimization \cite{ganopt2} and training performance as well as provide a guideline for choosing proper hyperparameters.
Third, we propose to employ the proposed model as an implicit proposal distribution in Bayesian state estimation to enhance tracking performance.
Finally, we apply the proposed approach to a task of vehicle behavior prediction.
 
The remainder of the paper is organized as follows. Section II provides a brief overview on related studies about GAN and Bayesian state estimation. Section III presents the details of proposed methodology. In Section IV, a numerical case study is demonstrated. In Section V, the proposed approach is applied to a real-world vehicle behavior prediction problem. Finally, Section VI concludes the paper.% and suggests future work.

\section{Related Work}
\noindent
\textbf{Generative Adversarial Network}

The core idea of training a deep generative model by making it compete with an opponent discriminator model was proposed in \cite{gan} where a two-player minimax game is formulated. A straightforward extension of incorporating condition information into both generator and discriminator was put forward in \cite{cgan} in which the generator can produce realistic images of particular categories.
There are mainly two aspects that can be modified to enhance the stability of GAN training: model setup \cite{dcgan,wgan,infogan} and optimization method \cite{ganopt1,ganopt3,ganopt2}. 
Although there have been a large number of studies on realistic image generation tasks  which provided promising results, much fewer research efforts, to our knowledge, have been devoted to continuous time-series data generation with adversarial training \cite{ts-gan1,ts-gan2}.
Our task is different from existing works where the generators are trained in unsupervised learning settings without any visualizable groundtruth distribution. This makes it hard to evaluate the learning performance. On the contrary, in this work the groundtruth actions can be utilized to monitor the training quality by calculating a proper error metric as well as provide a heuristic during the training process.

\noindent
\textbf{Multi-target Tracking and Prediction}

In recent years, widely studied multi-target tracking and prediction approaches can be mainly classified into two groups: 
(a) end-to-end solutions based on deep learning and computer vision techniques; (b) state estimation based on Bayesian inference.
In this paper, we only focus on the latter one with an emphasis on particle filter (aka. sequential Monte Carlo) due to its effectiveness and flexibility. 
\cite{PF_survey} provided a comprehensive summary of particle filter techniques. 
In the authors' previous work \cite{jiachen_tracking}, a generic multi-target tracking framework without explicit data association was proposed which can incorporate arbitrary learning-based state evolution models to enhance tracking performance. The framework can be applied to prediction problems as well provided that the measurement update is removed.
In this work, we employ the generator network in GAN as an implicit proposal distribution with high flexibility and generalizability which serves as the state evolution model in the tracking and prediction framework.

\section{Methodology}
In this section, we first make a brief introduction on the formulation and theory of GAN. Then, the proposed prediction model and algorithm are illustrated. Finally, the generic tracking framework is presented.

\subsection{Background and Preliminaries}
The Generative Adversarial Network is essentially a minimax two-player game including a generator $G$ and discriminator $D$. The goal of generator is to learn the data distribution and generate samples as similar to real data as possible while the goal of discriminator is to distinguish whether the given sample is real or produced by the generator. Additional condition information can be incorporated in this competing process. The value function of the minimax game is
\begin{equation}
\begin{aligned}
\min_G \max_D V(G,D) = \; &\mathbb{E}_{\mathbf{y} \sim p_{data}(\mathbf{y})}[\log D(\mathbf{y}|\mathbf{x})] \;+ \\
&\mathbb{E}_{\mathbf{z} \sim p_{\mathbf{z}}(\mathbf{z})}[\log (1-D(G(\mathbf{z}|\mathbf{x}))],
\end{aligned}
\end{equation}
where $\mathbf{z}$ is the latent noise sampled from normal distribution, $\mathbf{x}$ is condition information and $p_{\text{data}}(\mathbf{y})$ is the data distribution for generator to capture.
In practice, it is usually better for the generator to maximize $\log D(G(\mathbf{z}|\mathbf{x}))$ instead of minimizing $\log (1-D(G(\mathbf{z}|\mathbf{x})))$ due to the gradient saturation issue in the early training process.

\subsection{Model Architecture and Algorithms}
\begin{figure*}
\centering
\includegraphics[width=0.75\textwidth,height=5.5cm]{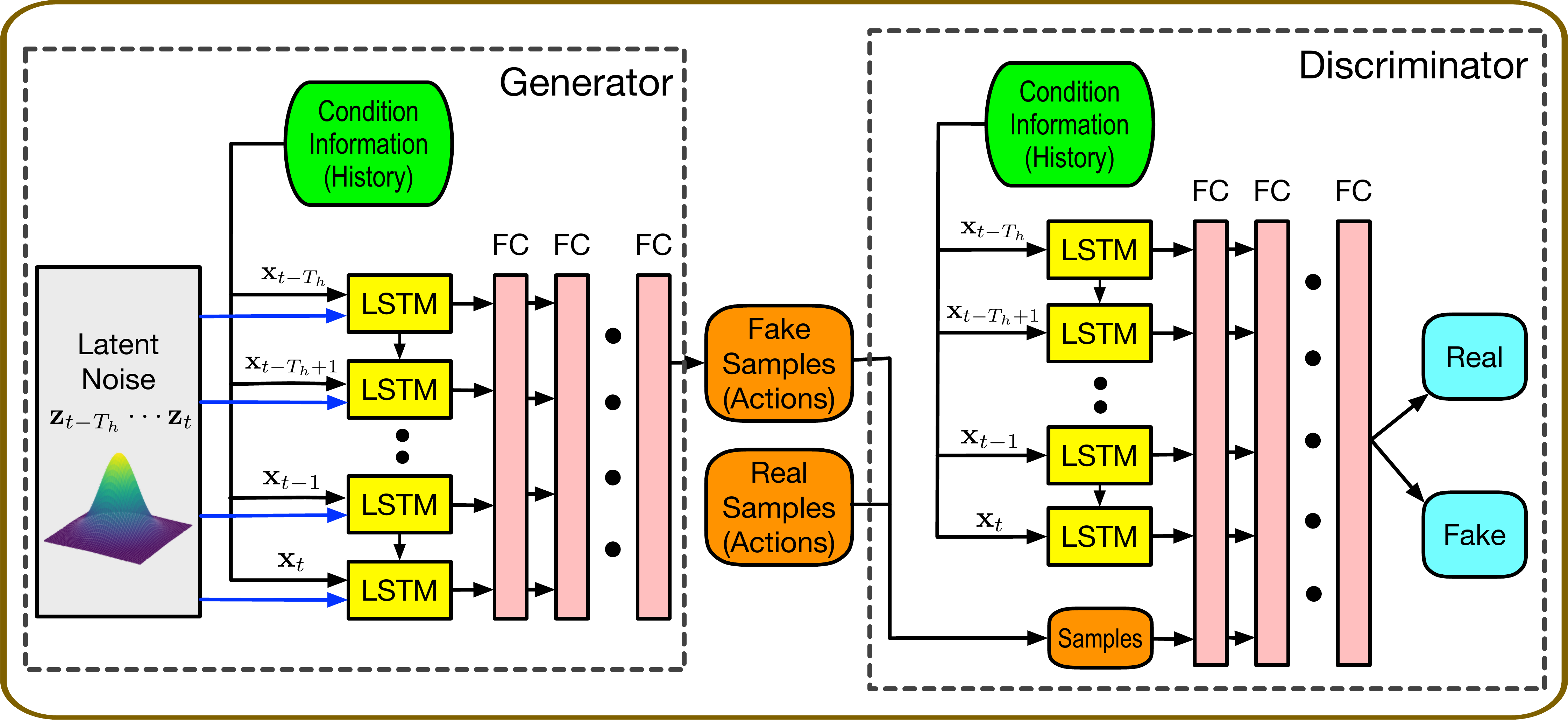}
\caption{The general diagram of the proposed model, which consists of a generator network and a discriminator network.}
\end{figure*}

The proposed model consists of a generator network and a discriminator network like the canonical GAN, which is shown in Fig. 1. The generator and discriminator have a similar architecture which is a combination of one recurrent layer and multiple fully connected (FC) layers. The Long Short-term Memory (LSTM) \cite{LSTM} is adopted due to its superiority on learning long-term dependencies as well as alleviating gradient vanishing problems.

As depicted in Fig. 1, the inputs of the generator network contain a sequence of latent noise $\mathbf{Z}=(\mathbf{z}_{t-T_h},\mathbf{z}_{t-T_h+1},...,\mathbf{z}_{t}), \; \mathbf{z}\in\mathbb{R}^{d}$ where $d$ is the dimension of noise vector and a sequence of condition information which refers to historical state trajectories $\mathbf{X}=(\mathbf{x}_{t-T_h}, \mathbf{x}_{t-T_h+1},...,\mathbf{x}_{t})$ in this work, although the model itself can be generalized to take in various forms of conditions. 
The noise vectors are sampled independently from a multivariate normal distribution $\mathcal{N}(\mathbf{0},\mathbf{I})$ or a uniform distribution $\mathbb{U}[0,1]$ at each time step. 
The outputs of the generator are fake samples which are actions of agents at the current time step in this work.
The discriminator network takes as inputs the same condition information as the generator, fake samples and real ones. They are concatenated with outputs of the LSTM layer. The probability of each sample being fake or real is provided by the discriminator output.

\noindent
\textit{\textbf{Training Phase:}}
Assume that GAN is a two players' zero-sum game, the loss function $V(G, D)$ is denoted as $f(\theta, \phi)$ where $\theta$ is the parameter vector of \textit{player 1} (i.e. discriminator) and $\phi$ is the parameter vector of \textit{player 2} (i.e. generator), then the objective functions are $f(x), -f(x)$ for the two players respectively. The optimum is a Nash equilibrium defined as:
\begin{equation}
\begin{aligned}
\theta^{*} \in \argmax_{\theta} f(\theta, \phi^{*}), \quad \phi^{*} \in \argmax_{\phi} -f(\theta^*, \phi).
\end{aligned}
\end{equation}
Define $x = (\theta, \phi)^{T}$, the Nash-equilibrium points possess the following properties:
\begin{equation}
\begin{aligned}
&v(x^*) =
\begin{bmatrix}
\nabla_{\theta}f(x^*)\\
-\nabla_{\phi}f(x^*)
\end{bmatrix}
=0,\\
H(x^*) =&
\begin{bmatrix}
\nabla_{\theta}^{2}f(x^*) & &\nabla_{\theta, \phi}f(x^*)\\
-\nabla_{\theta, \phi}f(x^*) & &-\nabla_{\phi}^{2}f(x^*)
\end{bmatrix}
\le 0,
\end{aligned}
\end{equation}
where $H(x)$ is the gradient of vector $v(x)$.

The standard optimization algorithm for GAN is simultaneous gradient ascent (SGA) which gives:
\begin{equation}
x_{k+1} = x_{k} + \alpha v(x),
\end{equation}
where $\alpha$ is the learning rate. The optimization is locally stable at Nash equilibrium if $H(x)$ is negative definite and $\alpha$ is small enough. However, in practice this naive update rule usually suffers a low covergence speed or even divergence, which makes the training process slow and unstable.
To avoid these issues, we use the consensus optimization proposed by \cite{ganopt2}. The objective functions of both players are added as a regularization term $-||v(x)||_{2}^{2}$. The update rule becomes: 
\begin{equation}
\begin{aligned}
x_{k+1} = x_{k} + \alpha(I - \gamma H(x)^{T})v(x),
\end{aligned}
\end{equation}
where $\gamma$ is the ratio between the objective functions $f(x), -f(x)$ and regularization term $-||v(x)||_{2}^{2}$. The optimization will be locally stable at the Nash equilibrium if $H(x)$ is invertible and $\alpha$ and $\gamma$ have proper values. 

This algorithm makes sure that the optimization procedure is locally convergent to the Nash equilibrium of the original game by selecting proper $\gamma$ and $\alpha$. 
Since $H(x)$ may have eigenvalues with zero real parts at Nash equilibrium which leads to limit cycles or divergence, this optimization scheme alleviates this issue by pushing the eigenvalues away from imaginary axis.
The generator and discriminator are trained end-to-end.

\noindent
\textit{\textbf{Testing Phase:}}
When we use the well-trained model to make predictions, only the generator network is employed to generate a group of future trajectory hypotheses given a certain historical state sequence. Since the generated samples only contain actions at current time step, they are processed to obtain the new state which is appended to the historical state sequence for propagation at the next step. Generating predictions of multiple time steps simultaneously is also a promising idea which is left to future research.

\subsection{Generic Mixture Tracking Framework}
The recursive Bayesian state estimation typically consists of two steps: prior update and measurement update. A state evolution model and a measurement model are required to obtain the state evolution distribution and measurement likelihood. 
The standard particle filter is able to provide a satisfactory approximation of the true posterior distribution in single object tracking problems; nevertheless, it usually suffers mode degeneration when the real distribution possess multiple modalities.
In order to maintain multi-modality arised from the existence of multiple tracking targets, the posterior state distribution is formulated as a mixture model 
\begin{eqnarray}
f(\mathbf{x}_k|\mathbf{z}^k) = \sum^{N}_{n=1}\pi_{n,k}f_n(\mathbf{x}_k|\mathbf{z}^k), \quad \ \sum^{N}_{n=1}\pi_{n,k}=1,
\end{eqnarray}
where $\mathbf{z}^k=(\mathbf{z}_0,...,\mathbf{z}_k)$, $N$ and $\pi_{n,k}$ refer to the component number of mixture distribution and corresponding weights, respectively. $f(\cdot)$, $\mathbf{x}_k$ and $\mathbf{z}_k$ refer to probability density function, state and measurement at the $k$th time step, respectively.
The canonical prior update and measurement update are modified into
\begin{equation}
\begin{aligned}
f(\mathbf{x}_k|\mathbf{z}^{k-1}) = &\sum_{n=1}^{N} \pi_{n,k-1} \iint [f_n(\mathbf{x}_k|\mathbf{x}_{k-1},\mathbf{e}_{k-1}, \mathbf{z}^{k-1})  \\
&\times f_{n}(\mathbf{x}_{k-1},\mathbf{e}_{k-1}|\mathbf{z}^{k-1})]\text{d}\mathbf{x}_{k-1}\text{d}\mathbf{e}_{k-1}, 
\end{aligned}
\end{equation}
\begin{equation}
f(\mathbf{x}_k|\mathbf{z}^{k}) = \frac{\sum_{n=1}^{N} \pi_{n,k-1} f_n(\mathbf{z}_k|\mathbf{x}_k) f_{n}(\mathbf{x}_k|\mathbf{z}^{k-1})}{\sum_{m=1}^{N} \pi_{m,k-1} \int f_m(\mathbf{z}_k|\mathbf{x}_k) f_{m}(\mathbf{x}_k|\mathbf{z}^{k-1})\text{d}\mathbf{x}_k},
\end{equation}
where $\mathbf{e}_{k-1}$ represents exterior information.
Please refer to \cite{jiachen_tracking} for more details about the tracking framework.

Since it is usually intractable and difficult to sample from the optimal proposal distributions $f_n(\mathbf{x}_k | \mathbf{x}_{k-1}, \mathbf{e}_{k-1}, \mathbf{z}_k)$, approximated ones $f_n(\mathbf{x}_k | \mathbf{x}_{k-1}, \mathbf{e}_{k-1})$ are usually employed to propagate particle hypotheses.
In this work, the generator of GAN essentially provides an implicit proposal distribution in the prior update.

\section{Numerical Case Study}
In order to demonstrate the performance of distribution learning of the proposed model, we present a numerical case study in which correlation and interaction exist among different random variables. 

\subsection{Problem Formulation}
The Lotka-Volterra system is a general framework of biological systems where the populations of predators and preys interact with each other which is a simple case suitable for simulating interactions among multiple agents as well as evaluating the distributions learned by generator.
The system equations are
\begin{equation}
\begin{aligned}
\dot{x} &= ax-bxy, \\
\dot{y} &= cxy - dy,
\end{aligned}
\end{equation}
where $a,b,c$ and $d$ are parameters.
The numerical dataset was generated by varying parameters and initial states. Parameters $a, b, c$ and $d$ were generated randomly from $\mathbb{U}[3, 5]$ and the initial states were generated randomly from $\mathbb{U}[1, 3]$.

\begin{figure*}[!tbp]
\centering
\includegraphics[width=\textwidth]{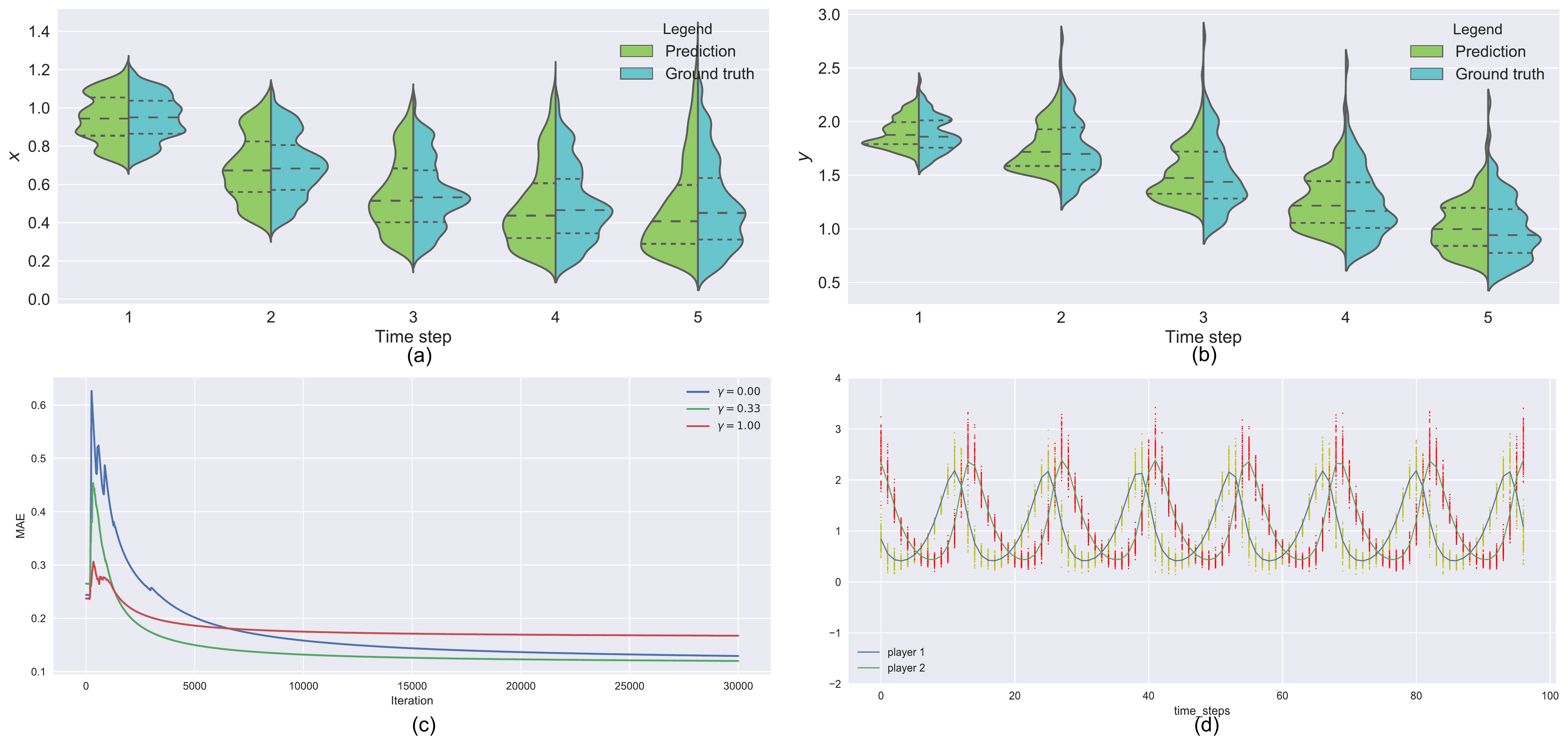}
\caption{Figures related to the numerical case study: (a) distribution comparison of $x$; (b) distribution comparison of $y$; (c) MAE values during the training process with different $\gamma$; (d) mixture tracking results.}
\end{figure*}

\subsection{Implementation Details}
The generator and discriminator network both consist of one LSTM layer with 128 units and four fully connected layers with 64 hidden units followed by a ReLU activation function except the final output layer. The dimension of latent noise is 16. Both networks were trained with RMSProp \cite{RMSprop} optimizer with a learning rate of 0.01. The hyperparameters were tuned manually with cross validation. We trained the model for 30,000 iterations with different regularization coefficients $\gamma=0.00,0.33,1.00$ to investigate the effects of regularization term. The standard loss function suggested in \cite{gan} was used in all the experiments.

\begin{figure}[!tbp]
	\centering
	\includegraphics[width=0.8\columnwidth]{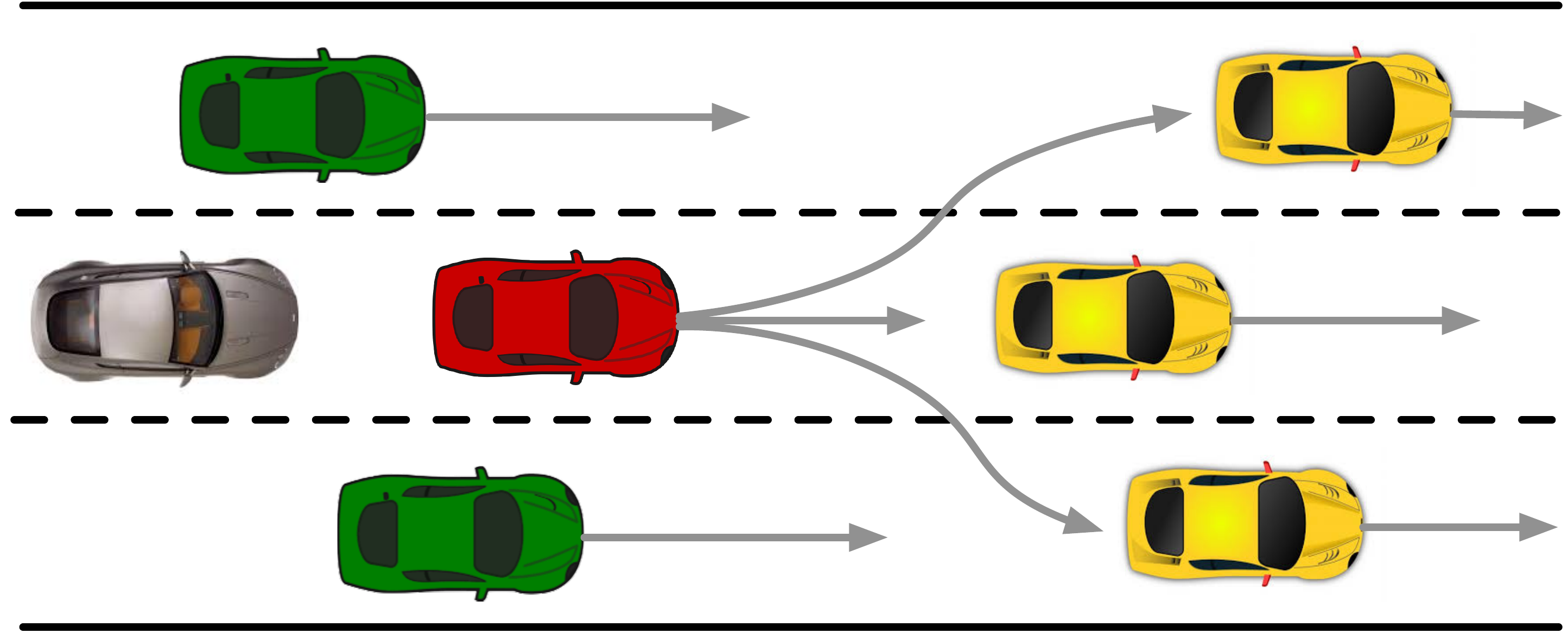}
	\caption{An illustrative diagram of highway scenario. The gray car is the ego vehicle which predicts future behaviors and motions of the other cars. }
\end{figure}

\subsection{Results and Discussion}

The training performance in terms of mean absolute error (MAE) values are shown in Fig. 2(c). It can be seen that higher $\gamma$ values may push the algorithm to a local optimum instead of the Nash equilibrium, which brings a trade off on the choice of regularization coefficient. 
In this numerical example, $\gamma=0.33$ leads to the best performance in terms of MAE during training process.

Since the objective of generator is to learn the data distribution, only employing the MAE metric is not sufficient. Therefore, we propose a generic distribution evaluation algorithm to obtain the groundtruth distribution of some statistic in the training dataset. 
For each given set of parameters $a,b,c$ and $d$, the groundtruth trajectories can be obtained by propagating the true model forwardly. Since the proposed model needs a sequence of historical information as input, we propagate the true model from the initial state backwardly to obtain historical trajectories which are employed to produce future trajectories using the proposed model.
More details are presented in \textbf{Algorithm 1}. 
We make a violin plot of the predicted distribution and the groundtruth distribution, which is illustrated in Fig. 2(a) and (b).
It is implied that not only can the generator capture the mean and variance of the real data distribution, but also it can make a close approximation to the shape of the groundtruth distribution in terms of multi-modality.

Moreover, the well-trained model is used as a probabilistic state evolution model in the mixture tracking framework to track the state of the predator-prey system. The tracking results with 100 particles are shown in Fig. 2(d). The mean of particles are close to the ground truth, and the variance is proper at each time step.

\begin{algorithm}[!tbp]
	\caption{Distribution Evaluation Algorithm}
	\begin{algorithmic}[1]
        \REQUIRE ~~\\
        	$para$: a set of different parameters $a,b,c$ and $d$;\\
    		$m$: number of different sets of parameters $a,b,c$ and $d$;\\
        	$n$: number of trajectories generated by the proposed model;\\
        	$T$: number of time steps to predict.\\
        \ENSURE ~~\\
        	$predicted\_traj$: trajectories generated by the proposed model;\\
        	$real\_traj$: trajectories generated by the true model.\\
		\FOR {$i = 1: m$}
        	\STATE $ s_{0:T} \leftarrow \text{generate\_forward\_trajectory}(para_{i})$\\
            \STATE $real\_traj_{i} \leftarrow s_{0:T}$\\
        	\FOR {$j = 1: n$}
            	\STATE  $s_{{-T_h}:{-1}}\leftarrow \text{generate\_backward\_trajectory}(s_{0}, para_i)$
                \STATE $\hat{s}_{-T_h:0} = s_{-k:0}$\\
            	\FOR {$t = 0 : T-1$}
                	\STATE $a_{t} \leftarrow \text{policy\_model}(\hat{s}_{t-T_h:t})$\\
                    \STATE $\hat{s}_{t+1} \leftarrow \text{update}(\hat{s}_{t},a_{t})$\\
                \ENDFOR
                \STATE $predicted\_traj_{ij} \leftarrow \hat{s}_{0:T}$
            \ENDFOR
        \ENDFOR
        \RETURN \text{histogram}($predicted\_traj, \ real\_traj$)
	\end{algorithmic}
\end{algorithm}

\section{Real-world Application: \\ Vehicle Trajectory Prediction}
In this section, we apply the proposed approach to solve a trajectory prediction task of interactive on-road vehicles as an illustrative example, although it can be utilized to solve many other tasks such as interactive pedestrian trajectory prediction and human-robot interactions.

\subsection{Problem Statement}

A typical highway scenario shown in Fig. 3 is investigated where the gray car is the ego vehicle which aims at forecasting future motions of its surrounding vehicles (red, green and yellow ones). The observations of environment can be obtained by on-board sensors. The approach can also be adopted in overhead traffic surveillance systems with camera-based monitors. 
We assume that only the red car can make a lane change to left (LCL) or right (LCR) while all the others maintain the lane keeping (LK) behavior. This is a reasonable assumption since it is rare in the realistic driving dataset that two or more vehicles change lanes simultaneously under the setup in Fig. 3. 
The behaviors of the red car and green cars are largely dependent on the their front yellow ones. Our objective is to learn a joint distribution of the six cars' future motions given a sequence of historical states to make long-term predictions.
We predict both the lateral deviation and longitudinal velocity of the red car at the next time step while only consider the longitudinal motions of surrounding cars since lateral positions within a lane make few effects during interaction as long as they are assumed to perform lane keeping behavior.

\subsection{Dataset and Preprocessing}
The NGSIM US-101 highway driving dataset \cite{NGSIM} was used to extract training data and test data with a ratio of 4:1. 
We selected 320 lane change cases and 2,800 lane keeping cases in total with a time horizon of 7 seconds for each case according to the requirements mentioned in Section V-A. 
Due to the existence of large detection noise on velocity and acceleration, an extended Kalman filter was applied to smooth the trajectories before feeding into the network.

\subsection{Experiments and Discussion}
The architectures of generator and discriminator network are identical to those in the numerical case study. The dimension of latent noise is 128. Both networks were trained with RMSProp optimizer with a learning rate of 0.0001. We trained the model for 100,000 iterations with a regularization coefficient of 0.1.
The state features include relative positions of five surrounding car with respect to the red middle car (set the position of red car as origin) and longitudinal velocities of all the cars. 
\begin{table*}[!tbp]
	\caption{MAE Comparisons of Vehicle Position Prediction (Center vehicle / Surrounding Vehicles)}
	\label{tab:feature}
	\begin{center}
		\begin{tabular}{m{2cm}<{\centering} m{2cm}<{\centering} m{2cm}<{\centering} m{2cm}<{\centering} m{2cm}<{\centering} m{2cm}<{\centering} m{2cm}<{\centering} m{2cm}<{\centering}}
			\toprule
			\midrule
			Cases &  Prediction Horizon (s) & Proposed model (m)& GMR (m)& P-MLP (m)& P-LSTM (m) & CAM (m)\\ % \hhline{=|=|=}
			\midrule
			\multirow{5}*{\shortstack[lb]{Lane Keeping \\(LK)}}  
			& 1.0 & \textbf{0.11}  /  \textbf{0.05} & 0.14 / 0.11 &0.27 / 0.27 &0.23 / 0.19 & 0.63 / 0.37\\ 
			& 2.0 & 0.83  /  \textbf{0.31} & 1.03 / 0.95 &\textbf{0.74} / 0.74 &0.82 / 0.59 & 1.93 / 1.28\\ 
			& 3.0 & \textbf{1.17}  /  \textbf{0.66} & 2.34 / 1.68 &1.20 / 1.42 &1.61 / 1.25 & 3.23 / 2.93\\ 
			& 4.0 & 2.31  /  \textbf{1.29} & 3.61 / 2.16 &\textbf{1.75} / 2.26 &2.69 / 1.91 & 4.73 / 3.88\\
			& 5.0 & 3.23  /  \textbf{2.53} & 4.18 / 3.86 &\textbf{2.97} / 3.39 &4.59 / 2.81 & 5.91 / 5.12\\
			\midrule
			\multirow{5}*{\shortstack[lb]{Lane Change \\(LC)}}  
			& 1.0 &  \textbf{0.08} / \textbf{0.04} & 0.21 / 0.13 &0.35 / 0.28 &0.34 / 0.25 & 0.51 / 0.44\\
			& 2.0 &  \textbf{0.46} / \textbf{0.26} & 0.93 / 1.06 &1.05 / 0.93 &0.96 / 0.83 & 1.56 / 1.47\\ 
			& 3.0 &  \textbf{1.21} / \textbf{0.71} & 2.13 / 1.74 &1.85 / 1.88 &1.69 / 1.57 & 2.68 / 2.11\\ 
			& 4.0 &  \textbf{1.97} / \textbf{1.47} & 3.26 / 2.48 &2.67 / 2.92 &2.91 / 2.69 & 4.72 / 3.64\\
			& 5.0 &  \textbf{3.12} / \textbf{2.71} & 5.21 / 4.06 &3.88 / 3.64 &4.00 / 3.78 & 7.33 / 5.27\\ 
			\bottomrule
		\end{tabular}
	\end{center}
\end{table*}
\begin{figure*}[!tbp]
	\centering
	\includegraphics[width=\textwidth]{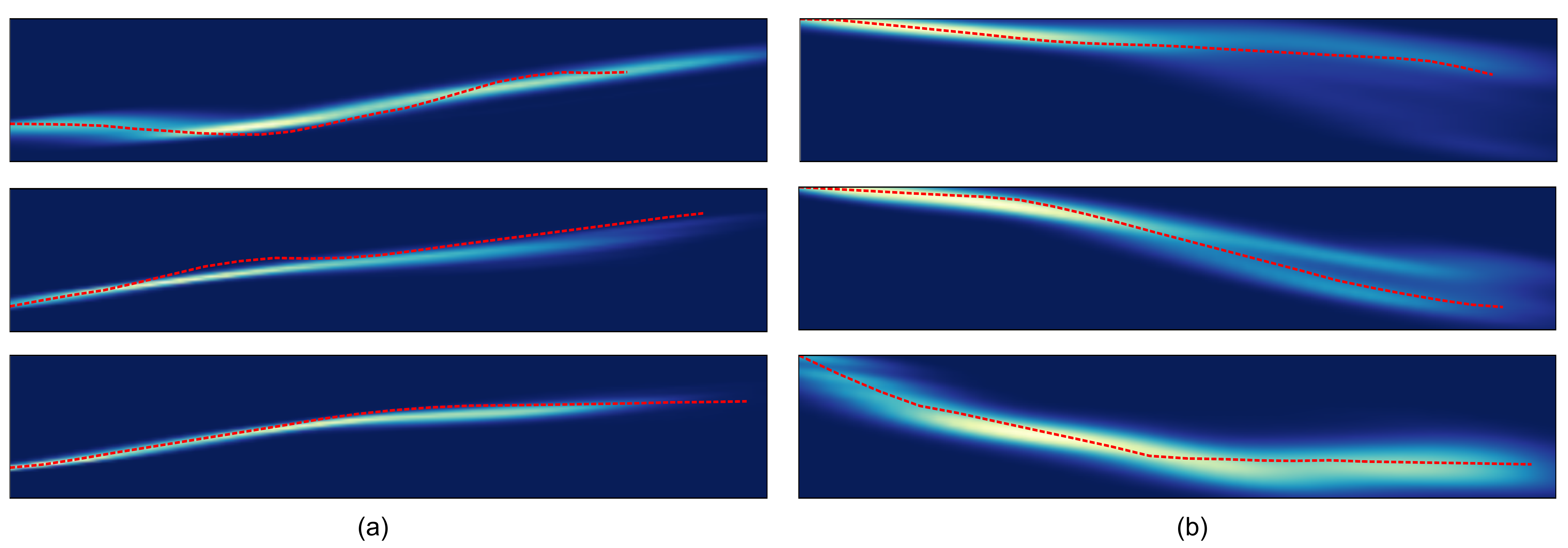}
	\caption{Visualization of selected cases. (a) lane change left cases; (b) lane change right cases. The red dash lines are groundtruth trajectories.}
\end{figure*}

We compared the prediction performance of the proposed model with the following baseline models using the same input state features and output actions:

1) \textit{Constant Velocity Model (CAM)}: CAM is one of the widely used kinematics models in vehicle tracking and prediction problems which assumes the vehicles go forward with a constant acceleration.

2) \textit{Gaussian Mixture Regression (GMR)}: This model is similar to the behavior model proposed in \cite{jiachen_tracking} in which the historical condition information and predicted actions are flattened and concatenated to fit a Gaussian mixture model in the training phase which gives the joint distributions of input and output. In the testing phase, the predicted actions can be sampled from the conditional distribution of actions given a certain conditional input.

3) \textit{Probabilistic Multi-layer Perceptron (P-MLP)}: In order to allow for a fair comparison, we added a Gaussian noise term to the input of MLP to incorporate uncertainty during both training and test process. We trained the P-MLP to minimize the mean square error (MSE) between the output actions and groundtruth actions. The network has five fully connected layers with 128 hidden units in each layer.

4) \textit{Probabilistic LSTM (P-LSTM)}: The P-LSTM is very similar to P-MLP except that the first hidden layer is replaced with an LSTM layer.

All the baseline models were trained with RMSProp optimizer for 100,000 iterations with a learning rate of 0.0001, which is identical to the training setup of the proposed model.

The MAE of prediction results of test cases are shown in Table I, in which the first number is the position error of middle vehicle while the second one is the mean longitudinal position error of surrounding vehicles. The reported values in the table are the average errors of multiple experiments. The bold numbers indicate best prediction performance under corresponding testing conditions.
It is shown that learning based models can achieve much better results than the CAM since the latter does not consider the interactions and uncertainty during the prediction period which suggests that vehicle kinematics models are only suitable for traffic scenarios with few interactions.
The proposed model can achieve the highest prediction accuracy in most cases except sometimes outperformed by P-MLP in lane keeping cases. The reason is that instead of minimizing the divergence of the generated sample distribution and groundtruth distribution, P-MLP tends to overfit training cases by maximizing likelihood which reduces model generalizability. Moreover, minimizing MSE loss may lead to average output of training action labels.
Therefore, although P-MLP works well in short-term prediction for lane keeping behaviors due to the low diversity of motion patterns, it fails to maintain a similar prediction accuracy for lane change behaviors where the variance of action distribution is much larger.
Also, it seems that the recurrent layer in P-LSTM has little help on improving prediction performance.
The GMR model can achieve better performance than P-MLP and P-LSTM in short-term prediction (less than 2 seconds) while it losses superiority as the prediction horizon increases.
Another issue is that we cannot incorporate arbitrary length of historical information in GMR since the data points become very sparse in high dimensional space which may lead to singular covariance matrix. This suggests that GMR is not suitable for learning long-term dependencies of time-series data.

Some prediction results of test cases are visualized in Fig. 4. It is illustrated that the groundtruth trajectories are located near the mean of the predicted distribution. Also note that in Fig. 4(b), the proposed model is able to forecast a multi-modal distribution and the groundtruth trajectory belongs to one of the modes.

\section{CONCLUSIONS}
In this paper, an adversarial learning based generic framework for interaction-aware multi-agent tracking and prediction was proposed, which can be generalized to any time-series prediction tasks. We also proposed to utilize the distribution learned by the generator as an implicit proposal distribution of mixture sequential Monte Carlo methods to improve state estimation accuracy.
A numerical case study was demonstrated to provide an empirical study on how to improve training performance of GAN through adding a regularization term to the standard loss function. The results show that increasing the regularization coefficient indeed accelerates convergence but may lead to sub-optimum which implies a trade-off in parameter tuning. It is also shown that our model can capture not only the mean and variance of the groundtruth data distribution but also the inherent multi-modalities. We also applied the proposed approach to a task of multi-vehicle trajectory prediction to illustrate its practicability and effectiveness.

%\addtolength{\textheight}{-12cm}   
%%%%%%%%%%%%%%%%%%%%%%%%%%%%%%%%%%%%%%%%%%%%%%%%%%%%%%%%%%%%%%%%%%%%%%%%%%%%%%%%
%\section*{APPENDIX}


%\section*{ACKNOWLEDGMENT}

%%%%%%%%%%%%%%%%%%%%%%%%%%%%%%%%%%%%%%%%%%%%%%%%%%%%%%%%%%%%%%%%%%%%%%%%%%%%%%%%
%\clearpage
\bibliographystyle{IEEEtran}
\bibliography{reference}

\end{document}